\documentclass[12pt]{iopart}
\usepackage{graphicx}
\usepackage{xurl}

\usepackage{xcolor}

\usepackage{natbib}

\begin{document}

\title[Mapping tree height from VHR images]{Sub-Meter Tree Height Mapping of California using Aerial Images and LiDAR-Informed U-Net Model} 

\author{F H Wagner$^{1,2}$, S Roberts$^{3}$, A L Ritz$^{4}$, G Carter$^{3}$, R Dalagnol$^{1,2}$, S Favrichon$^{2}$, M CM Hirye$^{5}$, Martin Brandt$^{3,6}$,  Philipe Ciais$^{3,7}$ and S Saatchi$^{1,2}$}


\address{$^1$ Institute of Environment and Sustainability, University of California, Los Angeles, CA, USA}
\address{$^2$ Jet Propulsion Laboratory, California Institute of Technology, 4800 Oak Grove, Pasadena, CA 91109, USA}
\address{$^3$ CTrees, Pasadena, CA 91105, US}
\address{$^4$ Virginia Polytechnic Institute and State University, Interdisciplinary
Graduate Education Program in Remote Sensing, Blacksburg, USA}
\address{$^5$ Quapá Lab, Faculty of Architecture and Urbanism, University of São Paulo, 05508080, São Paulo, SP, Brazil}
\address{$^6$ Department of Geosciences and Natural Resource Management, University of Copenhagen, Copenhagen, 1350, Denmark}
\address{$^7$ Laboratoire des Sciences du Climat et de l’Environnement, CEA-CNRS-UVSQ, CE Orme des Merisiers, Gif sur Yvette, 91190, France}

\ead{wagner.h.fabien@gmail.com}




\vspace{10pt}
\begin{indented}
\item[] \today 
\end{indented}

\begin{abstract}

Tree canopy height is one of the most important indicators of forest biomass, productivity, and species diversity, but it is challenging to measure accurately from the ground and from space. Here, we used a U-Net model adapted for regression to map the canopy height of all trees in the state of California with very high-resolution aerial imagery (60 cm) from the USDA-NAIP program. The U-Net model was trained using canopy height models computed from aerial LiDAR data as a reference, along with corresponding RGB-NIR NAIP images collected in 2020. We evaluated the performance of the deep-learning model using 42 independent 1 km$^2$ sites across various forest types and landscape variations in California. Our predictions of tree heights exhibited a mean error of 2.9 m and showed relatively low systematic bias across the entire range of tree heights present in California. In 2020, trees taller than 5 m covered $\sim$ 19.3\% of California. Our model successfully estimated canopy heights up to 50 m without saturation, outperforming existing canopy height products from global models. The approach we used allowed for the reconstruction of the three-dimensional structure of individual trees as observed from nadir-looking optical airborne imagery, suggesting a relatively robust estimation and mapping capability, even in the presence of image distortion. These findings demonstrate the potential of large-scale mapping and monitoring of tree height, as well as potential biomass estimation, using NAIP imagery.

\end{abstract}

%
%
%
%
%


\newpage
\section{Introduction}


California forests are particularly important among the world's forests as they are home to some of the tallest and largest tree species on Earth, such as coastal redwoods (\textit{Sequoia sempervirens}) and giant sequoias (\textit{Sequoiadendron giganteum}), respectively. They present some of the highest biomass in the world, reaching $\sim$ 2600 Mg of above-ground carbon per ha for \textit{Sequoia sempervirens} forests \citep{van2016}. Additionally, California forests exhibit high diversity and belong to one of the world's biodiversity hotspots, the California Floristic Province \citep{myers2000}. However, these forests are under threat. Studies have revealed a decline of 50 \% in the number of large trees in last decades \citep{mcintyre2015}, a recent decrease in carbon sequestration \citep{domke2020}, and an increasing vulnerability to disturbances caused by drought stress, insect infestation, and wildfires \citep{wang2022,Williams2016}. Consequently, urgent statewide efforts are being implemented to allow efficient forest conservation and management. These efforts require accurate information on the forest biomass and forest structure at the scale of meters for the entire California. While both values can be derived from canopy height \citep{asner2014m,lim2003lidar}, it introduces a challenging task for remote sensing techniques.

To estimate forest canopy height, airborne LiDAR (Light Detection and Ranging) is the gold standard, but data acquisition is limited to sparse, local to regional-scale coverage due to the high cost associated with it. To estimate tree height on a larger scale than the LiDAR flight lines, a novel approach is to estimate height measured from LiDAR with multispectral or radar remote-sensed images using machine/deep learning techniques. These models can then be applied to regions where LiDAR is not available. For example, estimations of tree height have already been made with Landsat images \citep{potapov2021m}, Sentinel-2 images \citep{lang2019, lang2022, astola2021}, combinations of Sentinel-1 (radar) and -2 (multispectral) \citep{ge2022, fayad2023vision}, Planet images \citep{Li20233, csillik2019, huang2022e}, and very high-resolution images from airborne \citep{li2020h, karatsiolis2021i} and satellite data \citep{illarionova2022e, tolan2023sub}.

Currently, two global vegetation height maps based on low and medium multispectral data are freely available: the 2020 Global Canopy Height Map at 10 m spatial resolution based on Sentinel-2 \citep{lang2022} and the 2019 Landsat-based global map at 30 m spatial resolution \citep{potapov2021m}. However, these maps, although useful for scientific applications, are not suitable for local forestry applications because of their coarser resolution and large uncertainty. Examples of tree height vegetation maps made with VHR optical imagery and deep-learning are increasing. A U-Net-based architecture was applied on VHR Worldview images over a site in boreal forests of Russia to map tree height with a mean absolute error (MAE) of 2.4 m for forests with average height of 15 m \citep{illarionova2022e}.  Similar models have achieved better estimates (1.4-1.6 m) for buildings and vegetation heights using aerial images \citep{li2020h}.  Others have  shown height estimation similar or greater accuracy (MAE $<$ 1.5 m) in urban and forest landscapes using encoder-decoder architecture on the 2018 Data Fusion Contest \citep{carvalho2019m, karatsiolis2021i, xu2019ad}. Overall, these studies show that estimation of height can be obtained with great accuracy using encoder-decoder deep learning architecture (such as U-Net \citep{Ronneberger2015}) for buildings or for vegetation from VHR images. However, these studies are all local proof of concepts and use a limited amount of training data, so it is still unknown how useful these models can be outside their training region.

Among the most recent similar studies, a deep learning-based framework was developed to provide location, crown area, and height for individual tree crowns from aerial images at a country scale \citep{Li20233}. It was trained in Denmark and successfully applied in both Denmark and Finland, demonstrating the potential for transferability to different countries. However, the approach involves separating the height estimation problem into two parts: segmenting individual tree crowns and attributing height and tree characteristics to the individual tree segments. Unfortunately, for California, some vegetation types, such as Chaparral, have continuous canopy cover, making it challenging to separate individual crowns in the image. Consequently, the first part of their model cannot be achieved for such vegetation types. The same issue, decomposing the model to segmentation and attributions, occurs with the recent canopy height model with a spatial resolution of 50 cm released by Meta \citep{tolan2023sub}. This model used the DINOv2 self-supervised model in conjunction with a deep learning model to estimate canopy height from Maxar RGB imagery with MAE of about 3-m over set-aside validation areas. Only training the DINOv2 model requires multiple high-end GPUs, which can also be a challenge for many research groups.

Here, we present a U-Net deep learning model adapted for regression which can directly predict vegetation height from the VHR image without decomposing the model to segmentation and attributions. It was trained using canopy height model (CHM) data from the USGS LiDAR campaign (2018-2020), the National Ecological Observatory Network (NEON) LiDAR campaign (2018-2020), and their corresponding VHR imagery from the NAIP USDA 2020 campaign. The model exclusively estimates vegetation height and differentiates it from other objects with height in LiDAR canopy height models, like buildings. The validation with 42 independent LiDAR datasets sampled across California are presented. Our canopy height estimates were compared to the available medium and very high-resolution canopy height models from Sentinel-2 (10 m), Landsat (30 m), and Worldview (50 cm). Finally, the California tree height map at 60 cm of spatial resolution is provided. 


\section{Methods}

  \begin{figure}[ht]
 \centering
 \includegraphics[width=1\linewidth]{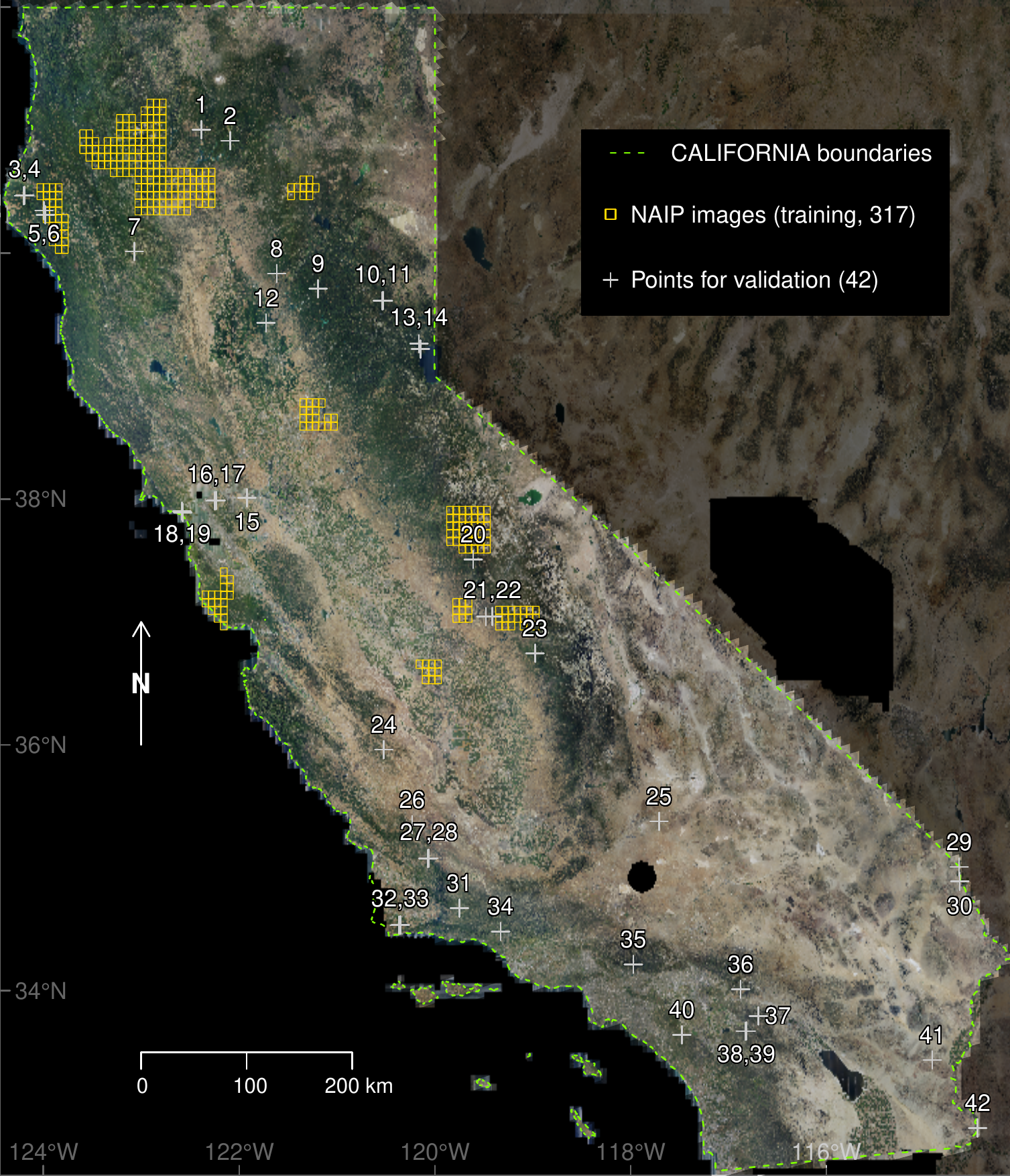}
  \caption{Mosaic of NAIP aerial images containing 11,076 tiles that cover the state of California, USA, from the 2020 campaign ($\sim$ 430,000 km$^2$). Only the red, blue, and green bands were used for visualization purposes. The map displays the extent of the NAIP images used for model training (shown in yellow) and the location used for model validation (shown with white numbers).}
  \label{Fig1}
  \end{figure}

 

\subsection{Very high resolution (VHR) optical images of California}\label{satimg}

To estimate the height of trees in California forests, a dataset of 11,076 aerial images covering the entire state of California from the US Department of Agriculture (USDA) National Agriculture Imagery Program (NAIP) of 2020 was used  \citep{NAIP2020}, Fig. \ref{Fig1}. Each NAIP image had a size of $\sim$ 6 $\times$ 7.5 km and a spatial resolution of 0.6 m. The images were acquired during the agricultural growing seasons between April and August 2020, using Leica Geosystems SH100 and SH120 sensors onboard an airplane. The provided bands included red, green, blue and  near infrared (RGBN). The NAIP images ($\sim$ 5 terabytes) were obtained from the USDA and delivered on a hard drive. The original images did not undergo any preprocessing as they were already within the range of 0 to 255 (8 bits).


\subsection{LiDAR dataset}


The LiDAR data to train the model were obtained from the U.S. Geological Survey (USGS) and the National Ecological Observatory Network (NEON) campaigns conducted in California forests between 2018 and 2020. Due to the extensive volume of LiDAR data available, we randomly selected specific tiles from each site. The model was trained using data from the following nine sites of the USGS  (\url{Lidar_CarrHirzDeltaFires}, \url{Lidar_LassenNP}, \url{Lidar_MountainPass}, \url{Lidar_SantaClaraCounty}, \url{Lidar_SantaCruzCo}, \url{Lidar_UpperSouthAmerican}, \url{Lidar_Yosemite}, \url{USGS_LPC_CA_FEMAR9Fresno_2019_D20}, \url{USGS_LPC_CA_NoCAL_3DEP_Supp_Funding_2018_D18}). In addition, from NEON, we used data from three sites: SOAP, SJER, and TEAK.

The model was validated using USGS LiDAR data from 2010 (one redwood site) and between 2018 and 2020 (10 sites). For the validation sample, first, 50 random points were generated in California, and the closest LiDAR point clouds were selected (18 LAZ files). Secondly, we generated 50 random points within forested areas as classified by the 2019 National Land Cover Database \citep{NLCD2019,yang2018new}. Again, the closest LiDAR point clouds to these points were selected (12 LAZ files). Additionally, we manually selected LiDAR data from forested regions, including areas with large and tall trees such as Redwoods and Sequoias (12 LAZ files). Therefore, these validation samples consisted of a total of 42 sites, which were used to assess the performance and accuracy of our model. The final list of validation data comes from 10 different USGS LiDAR datasets (\url{ARRA-CA_GoldenGate_2010}, \url{USGS_LPC_AZ_LowerColoradoRiver_2018_B18}, \url{USGS_LPC_CA_CarrHirzDeltaFires_2019_B19}, \url{USGS_LPC_CA_FEMAR9Estrella_2019_D20}, \url{USGS_LPC_CA_NoCAL_3DEP_Supp_Funding_2018_D18}, \url{USGS_LPC_CA_Riverside_2019_B19}, \url{USGS_LPC_CA_SE_Fault_Zone_Lidar_2017_D17}, \url{USGS_LPC_CA_SoCAL_Wildfires_2018_D18}, \url{USGS_LPC_CA_SouthernSierra_2020_B20}, and \url{USGS_LPC_CA_YosemiteNP_2019_D19}). 

Canopy height models (CHM) were generated for all the point clouds (originally in $.LAZ$ file format) using the \texttt{LidR} R package \citep{Roussel2020,Roussel2021}. First, the LiDAR point clouds were denoised to remove outliers using the $ivf$ algorithm with parameters of 1 m for resolution and 5 for the maximal number of other points in the surrounding \citep{Roussel2021}. Second, the digital terrain model (DTM) and the digital surface model (DSM) were computed at 1 m spatial resolution using the $TIN$ algorithm \citep{Roussel2021} and the $pitfree$ algorithm (thresholds of [0,2,5,10,15] and maximum edge length of [0, 1.5]), respectively. Third, the CHM was computed as the difference between DSM and DTM, multiplied by a factor of 2.5 and saved in integer 8 bits. Finally, the CHM data were resampled at 0.6 m using nearest neighbor algorithm, to match the original spatial resolution of the NAIP data. 

For the 42 independent validation sites, we compared the observed CHM and predicted CHM from our model and computed the mean average error (MAE), the MAE relative to the mean observed height, and the root mean square error (RMSE).





\subsection{Building footprints masking}

The 2020 US Building Footprints dataset made by Microsoft for the State of California was used to mask the heights of buildings that often appear erroneously in the CHM  (\url{https://github.com/microsoft/USBuildingFootprints}). A buffer of 2 m was added to the polygons of buildings to account for geolocation error, and the buffered polygons were rasterized to the CHM resolution. All CHM pixels overlapped by the buffered building footprints were set to zero. 

\subsection{Global tree height datasets}

The results of the model were compared to three recent global height datasets for our 42 validation sites. The first height dataset is taken from an unpublished global canopy height map for the year 2020 developed by Meta, Tolan's model \citep{metachm2023,tolan2023sub}. Currently, only the data for California (US) and the São Paulo State (Brazil) are available. This height map was produced at a 0.5 m spatial resolution using the self-supervised DINOv2 model and a machine learning approach that can estimate CHM from Maxar RGB satellite imagery while using aerial LiDAR from the NEON program in California as a reference to train the model \citep{metachm2023}. The second height dataset was obtained from a global CHM for the year 2020, Lang's model \citep{lang2022}. This height map was generated at a 10 m spatial resolution using CNN and Sentinel-2 reflectance data and also using GEDI LiDAR data as the reference for vegetation height. Currently, this dataset represents the most accurate freely available global vegetation height dataset. The third height dataset is taken from the 2019 Global Forest Canopy Height of the University of Maryland, Potapov's model \citep{potapov2021m}. In this dataset, the height is estimated at a 30 m spatial resolution from the Landsat normalized surface reflectance with a machine learning model (regression tree) using Global Ecosystem Dynamics Investigation (GEDI) LiDAR data as the reference for vegetation height \citep{potapov2021m,dubayah2020g,potapov2020l}. 






\subsection{Neural Network Architecture}

\begin{figure}[ht]
\centering
\includegraphics[width=1\linewidth]{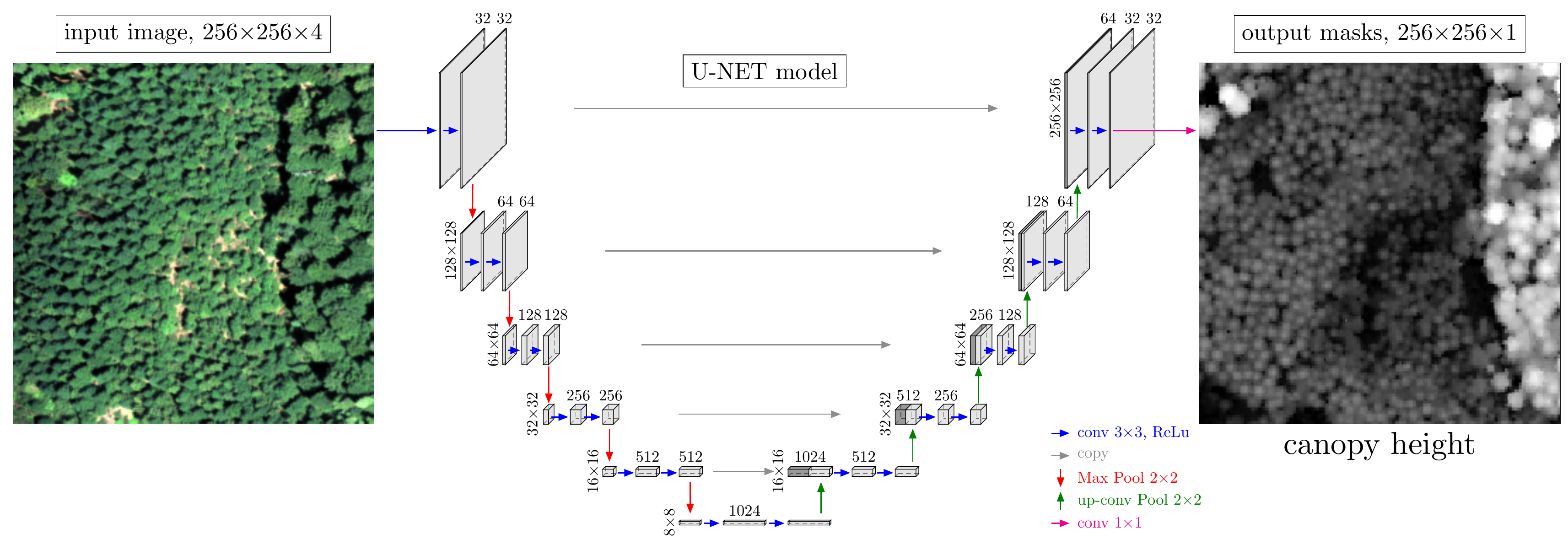}
 \caption{U-Net model architecture used for canopy height estimation from VHR NAIP images.}
 \label{FigUnet} 
 \end{figure}

The canopy height estimation from the NAIP images of California was performed using a classical U-Net model \citep{Ronneberger2015} with $\sim$ 35 millions parameters, as depicted in Fig. \ref{FigUnet}. Specifically, the U-Net model predicted the canopy height for each pixel of the input image. The model input was a 4-band RGB-NIR image with dimensions of 256 $\times$ 256 pixels and a spatial resolution 0.6 m. The output of the model was a single-band mask with dimensions of 256 $\times$ 256 pixels, containing values ranging from 0 to 1 (representing 0 to 100 m when unscaled). The model was implemented using the R language \citep{CoreTeam2016} with the RStudio interface to Keras and TensorFlow 2.8 \citep{chollet2015keras,AllaireChollet,allaireTang,AbadiAgarwalBarhamEtAl2015}. 



\subsection{Training}

To generate the training samples, we initially selected 7,956 CHM rasters, out of the original 11,225 CHM rasters, that were completely covered by a NAIP image. Then NAIP images (317) were cropped to match the extent of each CHM. Subsequently, we resampled the 1 m spatial resolution CHM rasters to the resolution of the cropped NAIP image (0.6 m) using the nearest neighbor algorithm. In the third step, both the cropped NAIP and corresponding CHM rasters were divided into image patches of 256 $\times$ 256 pixel cells using the \texttt{gdal\_retile} tool \citep{gdal2019}. Any image patch that deviated from the size of 256 $\times$ 256 pixels was excluded, resulting in a total of 145,080 image patches. Additionally, to train the model to recognize when there was no vegetation height, we incorporated 15,310 image patches of 256 $\times$ 256 pixels without vegetation (e.g., deserts, rocks) or only water surfaces and their CHM rasters consisting solely of zeros.



The final training sample comprised a total of 160,390 256 $\times$ 256 pixels image patches of 4 bands and their associated one-band CHM image patch. 90\% (144,351) of the samples were used for training and 10\% (16,039) for validation. Before being inputted into the U-Net model, image patches underwent random vertical and horizontal flips as data augmentation.


 
During the network training, we used standard stochastic gradient descent optimization and the RMSprop optimizer (learning rate of 0.0001). Mean squared error was used as the loss function, and mean absolute error as the accuracy metric. The network was trained for 5,000 epochs with a batch size of 32 images. The model with the best validation loss (i.e., validation loss of 0.001954955 and a mean absolute error of 2.30 m) was kept for prediction. The training of the models took less than a week using an Nvidia RTX3090 Graphics Processing Unit (GPU) with 24 GB of memory.




  
 \subsection{Prediction}

For prediction, NAIP tiles were expanded by adding columns and rows, resulting in images of 11,392 $\times$ 13,440 pixels with a proportional aspect ratio of 1024 and a 128-pixel border. These images were then divided into sub-images of 1,152 $\times$ 1,152 pixels with a 64-pixel overlap to mitigate border artifacts \citep{Ronneberger2015}. The prediction was performed on each sub-image, cropped to a slightly smaller extent (removing the 64 pixels margin), and then merged to obtain the NAIP tile of predicted height. The prediction time for California was 23 days using the RTX3090 GPU.

\section{Results}

\subsection{Validation of the model using 42 independent sites}

\begin{figure}[ht]
\centering
\includegraphics[width=0.78\linewidth]{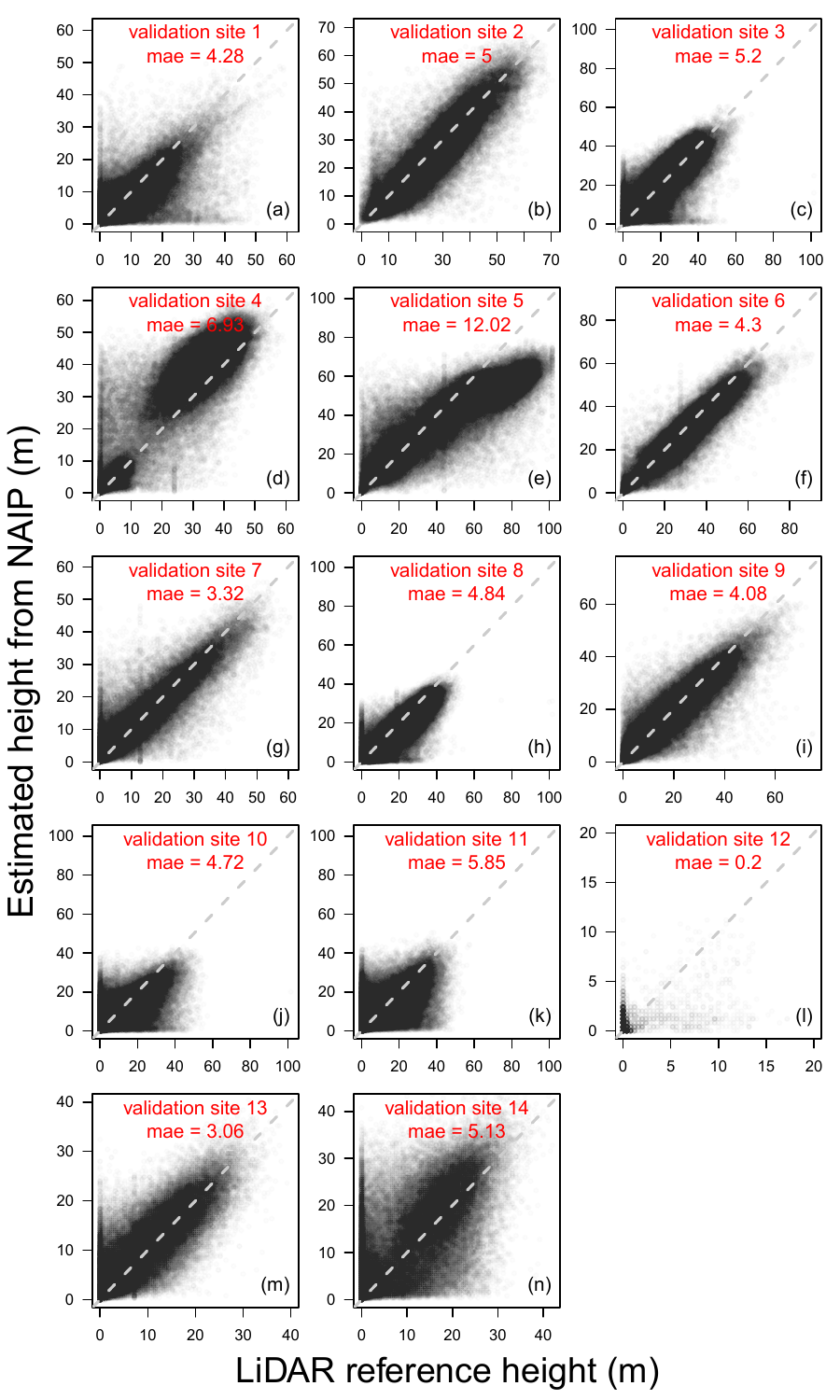}
 \caption{Comparison of predicted versus observed height (m) for the validation sites in the Northern part of California (Panel 1). Each plot consists of 100,000 points, $\sim$ one-tenth of the total observed points. The 1:1 line is depicted in gray, and the mean average error is indicated in red.}
 \label{Figpredobs1} 
 \end{figure}

 The model accurately predicted the height, as evidenced by the alignment of predictions and observations along the 1:1 line, for most of the 42 validation sites across the range of canopy heights in California, from North to South (Fig. \ref{Figpredobs1}at the top right to Fig. \ref{Figpredobs3} at the bottom center). It is important to note that the NAIP and CHM images are not co-registered, and the trees in the NAIP images may appear distorted due to variations in view angle and slope. 

\begin{figure}[ht]
\centering
\includegraphics[width=0.78\linewidth]{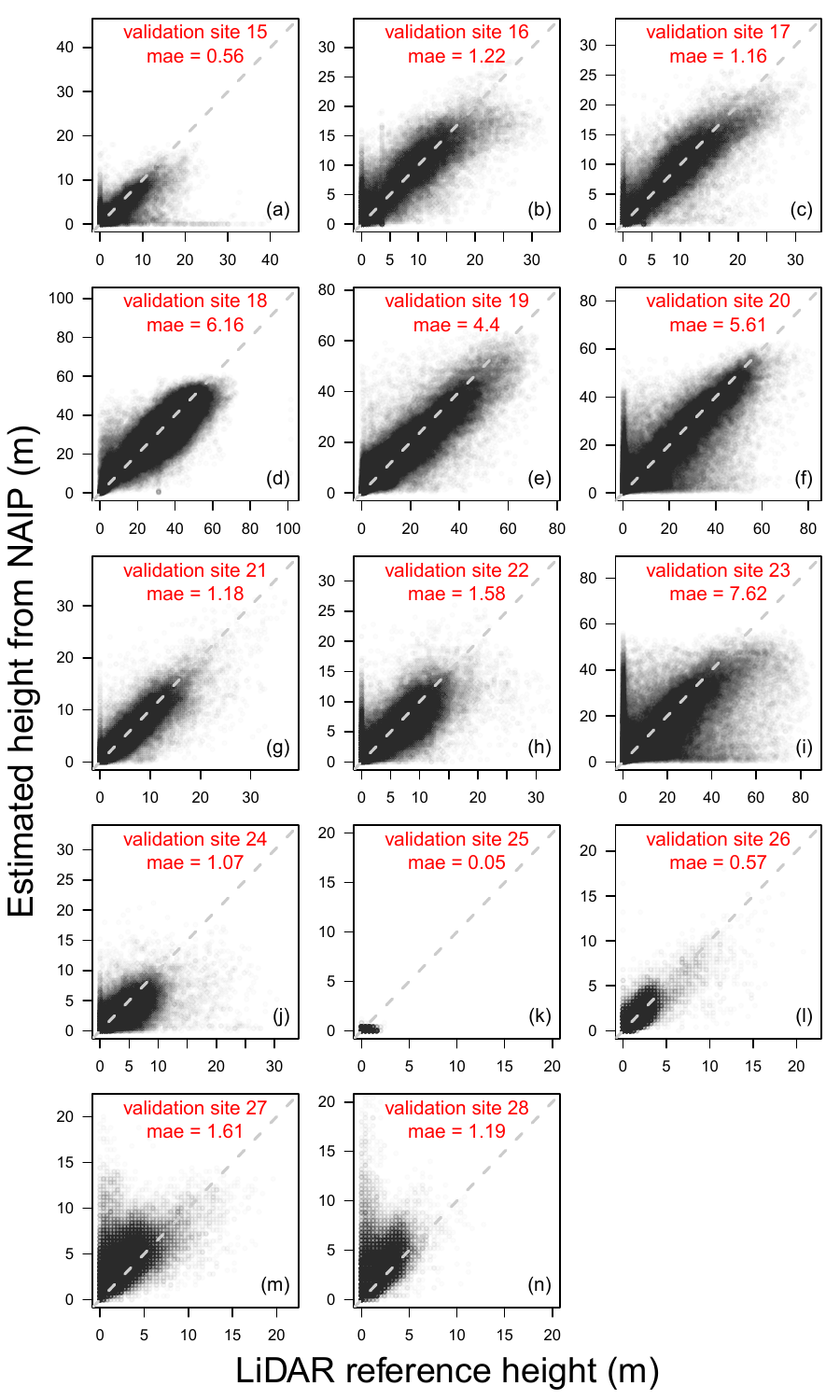}
 \caption{Comparison of predicted versus observed height (m) for the validation sites in the central part of California (Panel 2). Each plot consists of 100,000 points, $\sim$ one-tenth of the total observed points. The 1:1 line is depicted in gray, and the mean average error is indicated in red.}
 \label{Figpredobs2} 
 \end{figure}

\begin{figure}[ht]
\centering
\includegraphics[width=0.78\linewidth]{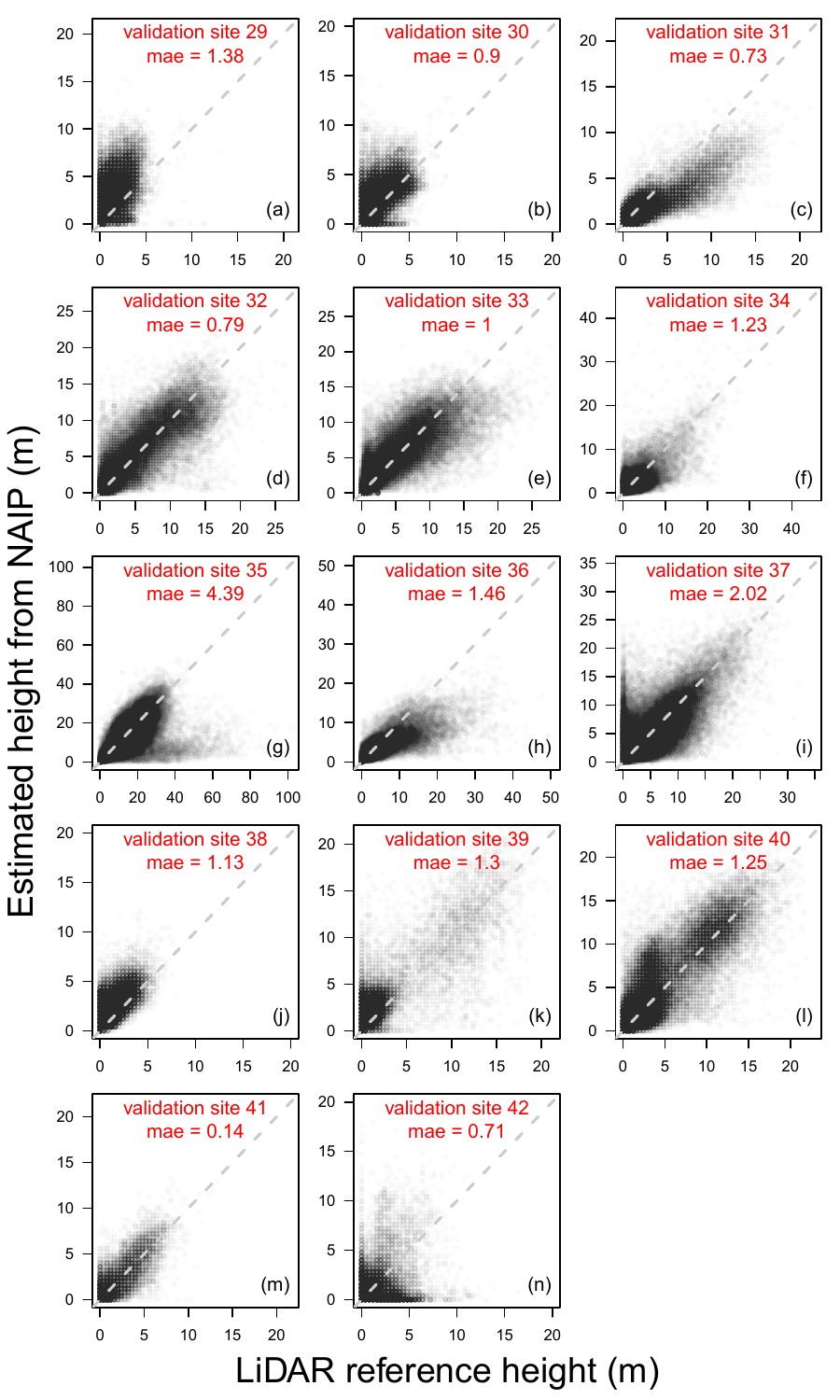}
 \caption{Comparison of predicted versus observed height (m) for the validation sites in the Southern part of California (Panel 3). Each plot consists of 100,000 points, $\sim$ one-tenth of the total observed points. The 1:1 line is depicted in gray, and the mean average error is indicated in red.}
 \label{Figpredobs3} 
 \end{figure}

For most sites with tall forests reaching over 60 m (Fig. \ref{Figpredobs1}b, c, d, e, f, g, and i, and Fig. \ref{Figpredobs2}d, e, f, and i),  the model can accurately predict heights up to 50 m without bias. Between 40-50 m, there may be a slight underestimation, but it does not reach a saturation, as seen in the Redwood forest (Fig. \ref{Figpredobs1}e) or the Sequoia forest (Fig. \ref{Figpredobs2}i). However, underestimation is not always observed, Fig. \ref{Figpredobs1}b, f and Fig. \ref{Figpredobs2}e, f. In the Redwood forest (Fig. \ref{Figpredobs1}e), the tall trees are densely packed, making it visually challenging to distinguish between 40 m and 70 m trees from the NAIP image alone (Fig. \ref{val05}).


The model accurately identifies most trees, with only a few ground points being misclassified. The misclassification of ground points as vegetation height occurs primarily in tall forests and is concentrated near zero on the observed height axis (Fig. \ref{Figpredobs1}g, h, j-n, Fig. \ref{Figpredobs2}f and i, and Fig. \ref{Figpredobs3}i). This error can be attributed to a slight discrepancy in the location of observed and predicted large and tall crowns, as illustrated in Fig. \ref{val20} and Fig. \ref{val23}. When this occurs, the overlapping portion of the crowns between the predicted and observed values shows a small error, whereas the border where the observed and predicted crowns do not overlap displays high negative errors on one side and positive errors on the other side.


Outliers that are predicted as zero but observed with a height, Fig. \ref{Figpredobs2}a and k, and Fig. \ref{Figpredobs3}n, mainly result from buildings or electric lines that are mistakenly classified as vegetation in the USGS point clouds and consequently appeared in the reference CHM. However, the model correctly predicts a height of zero for them, which is the accurate value (see Fig. \ref{val15} and Fig. \ref{val42}). It should be noted that our model may not detect vegetation below two meters in height, as observed in the desert (Fig. \ref{val25}). Furthermore, fire or logging activities may have resulted in the removal of some trees between the acquisition of the NAIP image and the LiDAR point cloud, as seen in validation site 1 (Fig. \ref{val01}) and site 10 (Fig. \ref{val10}).





\clearpage

\clearpage
\newpage

 \clearpage
\newpage

 \begin{figure}[ht]
\centering
\includegraphics[width=0.5\linewidth]{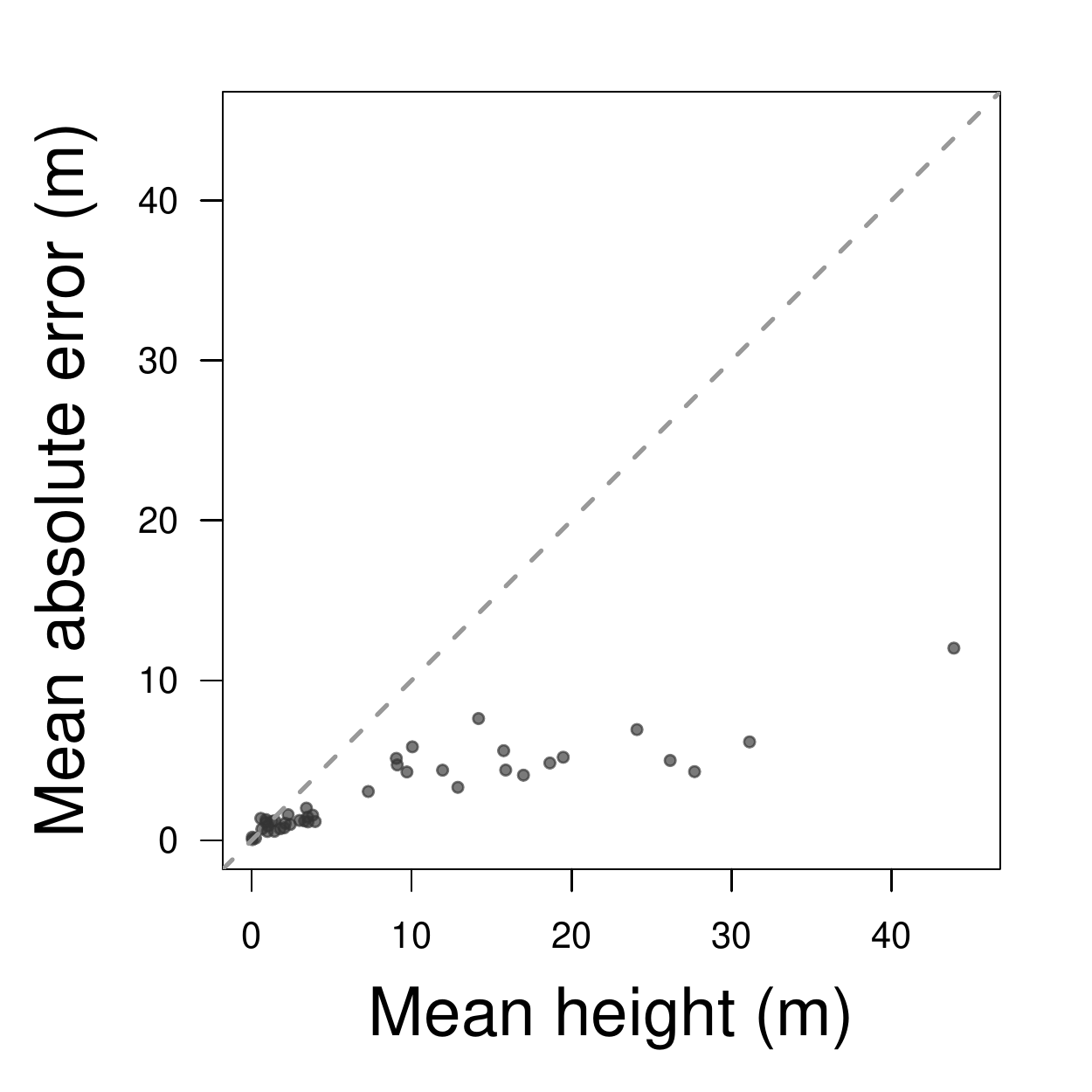}
 \caption{Association between the mean observed canopy height and the mean average error across the 42 validation sites. The 1:1 line is depicted in gray.}
 \label{Figmae} 
 \end{figure}

The mean average error (MAE) of our model on the validation sample was 2.90 m. MAE ranges from 0.05 m in a desert area with very low vegetation and few trees to 12.02 m in the tallest observed redwood forest, Fig. \ref{Figmae}. The MAE increased linearly with the mean tree height, Fig. \ref{Figmae}. This suggests that certain features used by the model to reconstruct the CHM from the VHR images are less accessible for taller and denser forests. Furthermore, if the prediction is good but has a small location error, this could also explain the increase in MAE with an increase in tree height. The largest discrepancies between the reference and the predicted heights are generally observed on the border of the highest tree crowns Fig. \ref{val23}). This geolocation artifact is likely responsible for the peak of predicted height often observed for low values of observed height, such as in Fig. \ref{Figpredobs1}-\ref{Figpredobs3}.






\subsection{From a 2D multispectral image to a 3D canopy height model}

\begin{figure}[ht]
\centering
\includegraphics[width=1\linewidth]{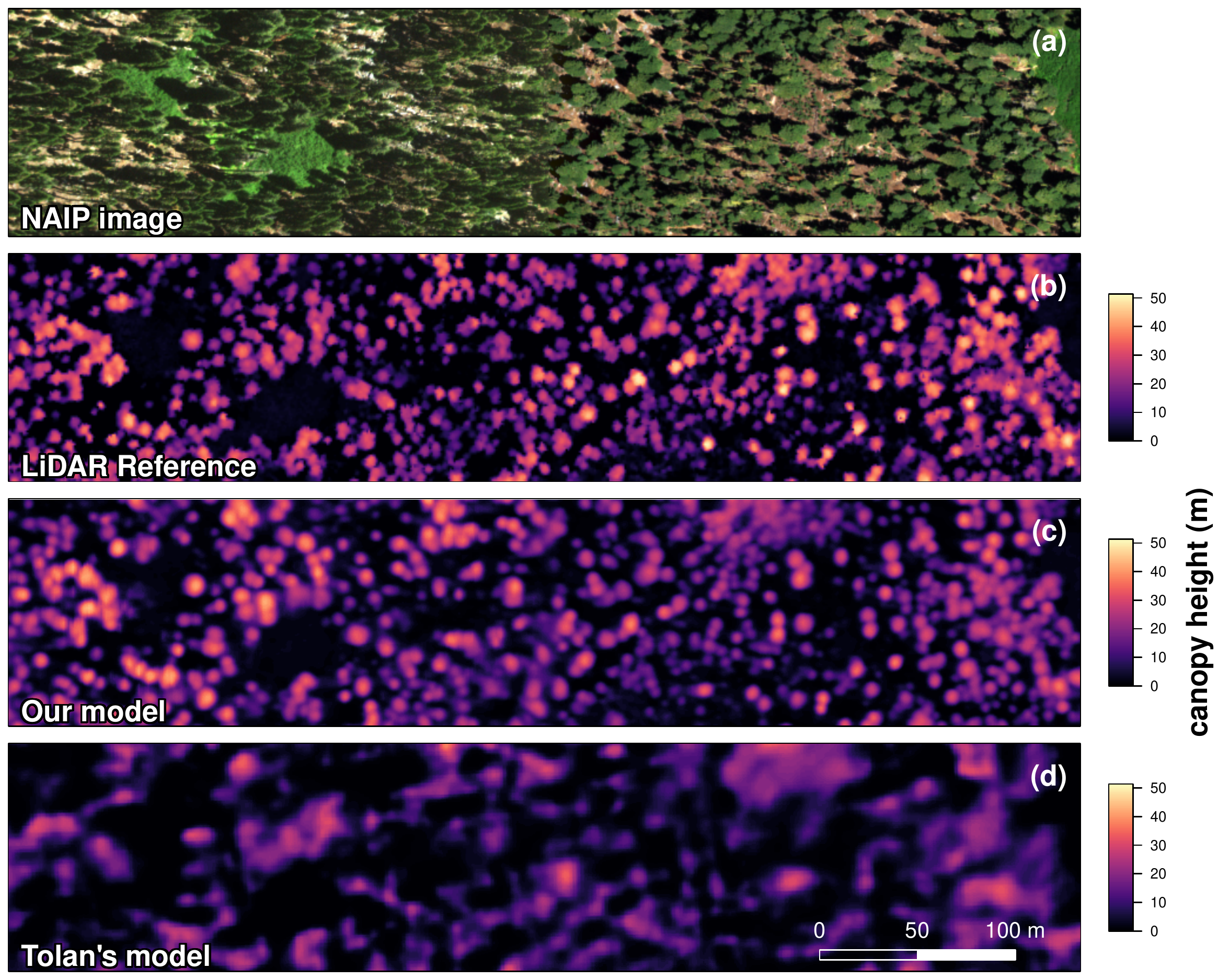}
 \caption{Example of a NAIP image from the validation sample displaying flattened tree artifacts due to the viewing angle (a), reference canopy height model obtained from LiDAR (b), canopy height model generated by our model (c), and canopy height model produced by Tolan's model \citep{tolan2023sub} (d).}
 \label{2D3D} 
 \end{figure}

Surprisingly, our model reconstructed the 3D structure of trees from the multispectral NAIP image, Fig. \ref{2D3D}.  It generated a CHM from a nadir view, similar to the LiDAR reference data, despite the NAIP sensor’s view angle not being exactly nadir and the presence of different view angles in the image mosaic resulting from various image acquisitions, as observed in validation site 11 (Fig. \ref{2D3D} and Fig. \ref{val11}). Although the trees may appear distorted and flattened in the NAIP image (Fig. \ref{2D3D}a), and the LiDAR reference appears visually different from the NAIP image (Fig. \ref{2D3D}b), our model successfully reconstructed a realistic CHM, including tree positions and crown sizes. This reconstructed information can be accessed at the individual tree level (Fig. \ref{2D3D}c). However, due to slight differences in tree location between the predicted and observed data, significant negative and positive errors occur at the borders of large crowns (Fig. \ref{val11}). When compared to Tolan's CHM (Fig. \ref{2D3D}d), which is generated using VHR satellite images, it can be observed that Tolan's model did not recreate the 3D structure of trees at the individual tree level. Instead, it estimated heights at a lower resolution than the original data, and information on individual trees could not be accessed.



\subsection{Comparison with global height products for the 42 validation site}
\clearpage
 \begin{figure}[ht]
\centering
\includegraphics[width=0.7\linewidth]{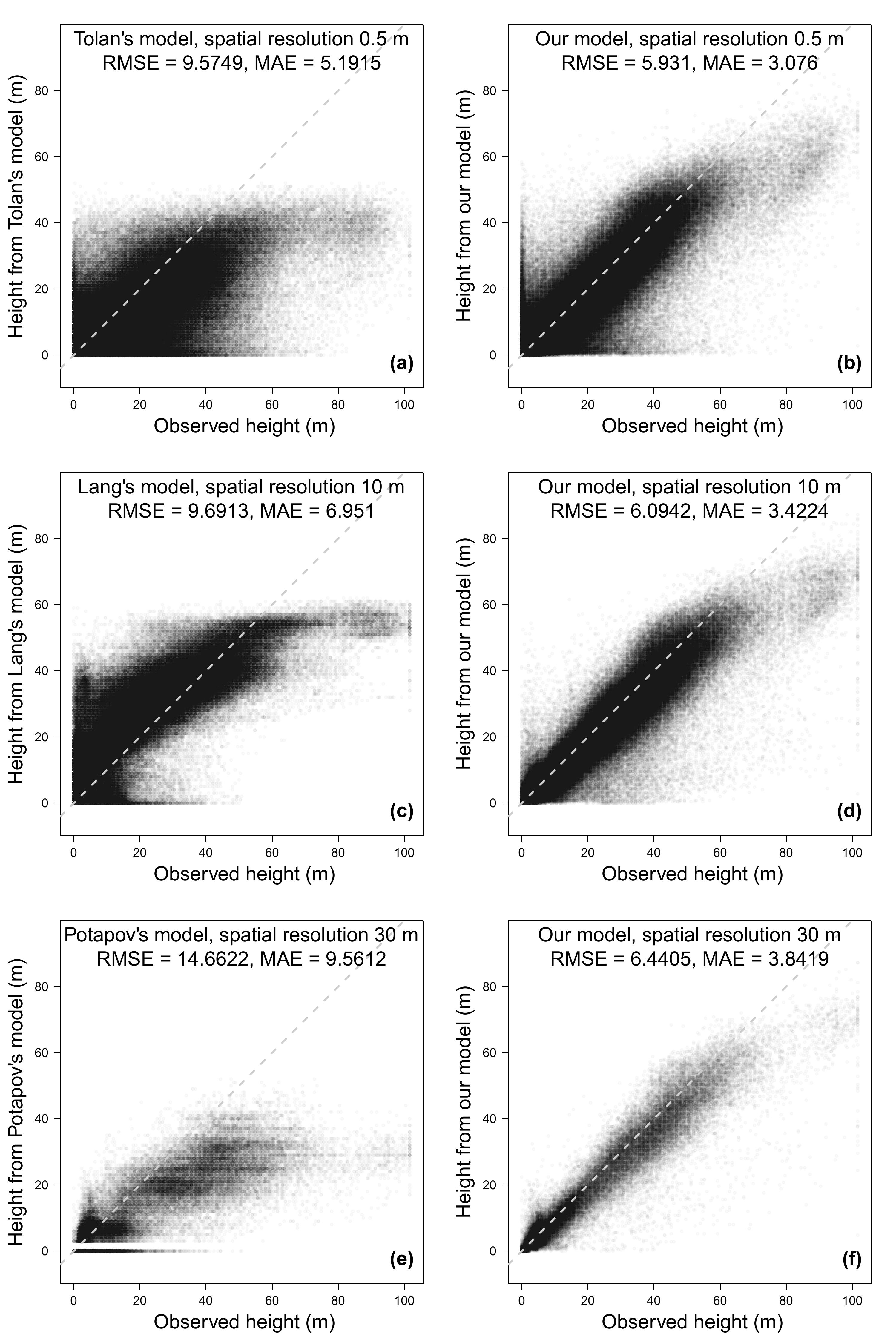}
 \caption{Comparison with global height products for the 42 validation sites in California. (a) Predicted versus observed height (m) for Tolan's 50 cm spatial resolution canopy height model \citep{tolan2023sub};  (b) Predicted versus observed height of our model aggregated at the 50 cm spatial resolution of Tolan's model; (c) Predicted versus observed height (in meters) for the 10 m spatial resolution canopy height model of Lang \citep{lang2022};  (d) Predicted versus observed height of our model aggregated at the 10 m spatial resolution of Lang's model; (e) Predicted versus observed height (in meters) for the 30 m spatial resolution canopy height of Potapov \citep{potapov2021m}; (f) Predicted versus observed height of our model aggregated at the 30 m spatial resolution of Potapov's model.  For (a) and (b), there are 183,947,110 points of validation; for (c) and (d), there are 525,640 points; and for (e) and (f), there are 65,633 points. The 1:1 line is represented in gray.}
 \label{FigGlob} 
 \end{figure}




Tolan's model \citep{tolan2023sub}, made using VHR satellite images at a 0.5 m resolution,  tends to consistently underestimate vegetation height and saturates before 45 m height (Fig. \ref{FigGlob}a).  The model’s error is not uniform across different heights, with larger errors occurring for smaller heights, and the height estimation reaches a threshold at 40 m. In contrast, our model is capable of predicting greater heights (Fig. \ref{FigGlob}b). The error remains constant for all heights below 50 m, and beyond 50 m, there is an observed underestimation of height. The large errors for points observed with an elevation of zero are associated with predicted crown location errors. In comparison to Tolan's model results, the mean average error (MAE) and root mean square error (RMSE) of our model are $\sim$ 1.7 times and 1.6 times lower, respectively, and our model estimates saturate at a much taller height, $\sim$ 75 m.


Lang's model \citep{lang2022} roughly follows the 1:1 line  (Fig. \ref{FigGlob}c), but the relationship to the airborne LiDAR is not linear. This is caused by the custom weighting included in their model in order to alleviate the saturation effect in high vegetation heights. The mean average error (MAE) of the Sentinel 2-based model is twice as large (MAE = 6.95 m) as that of our locally trained model (MAE = 3.4 m). Additionally, this dataset reached greater heights than Tolan's model, saturating close to 60 m. At the spatial resolution of Sentinel-2 (Fig. \ref{FigGlob}d), our model appears to be closer to the 1:1 line, and the peak of error observed at an observed height of zero disappears. This suggests that a portion of the errors in our model is not related to height estimation but rather to tree location inaccuracies. When aggregated at a larger spatial resolution, these errors vanish as most of these significant discrepancies are located at the borders of the crowns.

Potapov's model exhibits underestimation of vegetation height for the validation sites, reaching a threshold around 40 m (Fig. \ref{FigGlob}e), as already described in the original article of the model \citep{potapov2021m}. At the spatial resolution of 30 m, our model’s results appeared to be closer to the 1:1 line (Fig. \ref{FigGlob}f), at least up to 50 m. Beyond that point, our model shows an underestimation trend until reaching saturation around 75 m.


\subsection{California canopy height}

\begin{figure}[ht]
\includegraphics[width=1\linewidth]{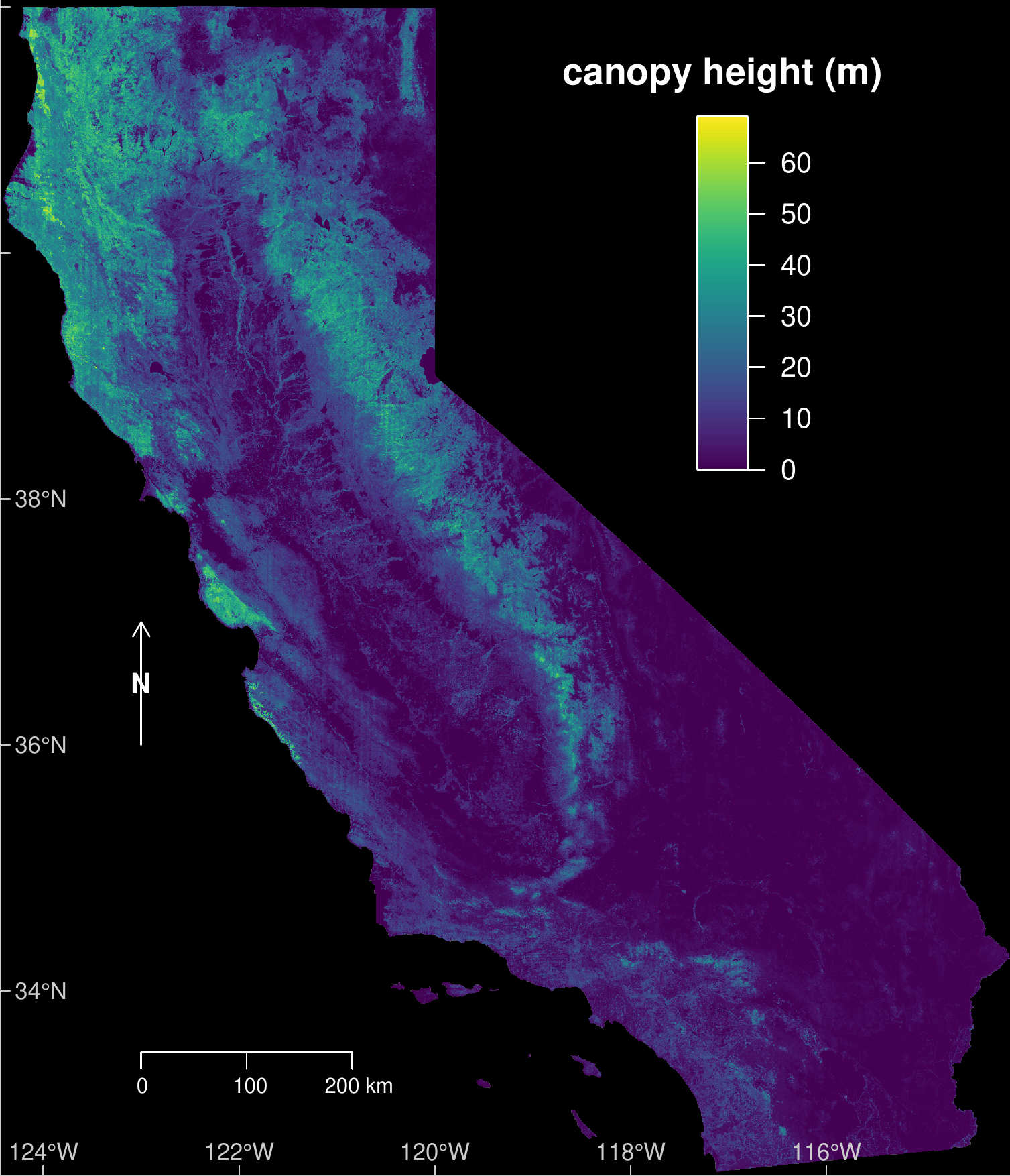}
   \caption{Canopy height of California (m). To ease visualization, colors represent the estimations from our model aggregated at a 30 m spatial resolution using the median.} 
  \label{CaliHeight}
  \end{figure}

The following statistics on vegetation height are provided based on the median canopy height aggregated at a 30 m spatial resolution (Fig. \ref{CaliHeight}). In 2020, we determined that 31.9\% of California had vegetation heights $\geq$ 2 m. By using a median height threshold of 5 m to define forests, we estimated that forest coverage accounted for $\sim$ 19.3\% of California. The median forest height in California was 11 m, with 5\% of the forests exhibiting a median height above 29 m. Among the forests, 0.7\% had a median height of 40 m or higher, and 0.07\% had a median height of 50 m or higher. The maximum observed median height at the 30 m resolution was 72 m, likely close to the saturation point of our model. The tallest forests were primarily distributed in the Klamath Mountains, Cascade Ranges, Moloch Plateau Ranges, Sierra Nevada Mountains, northern part of the Coast Ranges, Santa Cruz Mountains, and Santa Lucia Range.


\section{Discussion}

\subsection{Mapping California tree height}

Here, we demonstrate that very high spatial resolution optical aerial images, such as NAIP, allow direct measurement of forest canopy height. In California, the U-Net network adapted for regression estimated canopy height directly from VHR NAIP images with a mean error of 2.9 m on the independent validation datasets, demonstrating again the high capacity of convolutional networks to support vegetation mapping \citep{KATTENBORN202124}. Overall, for California, our model provided better canopy height estimates with fewer biases than the currently available global canopy height models (Fig. \ref{FigGlob}). Considering that California is a biodiversity hotspot \citep{myers2000} and exhibits a wide range of trees and forest structures, from sparse vegetation in the desert to the tallest trees on earth, it is likely that this type of model will work for most temperate forests  although further testing is required. 

We found that 19.3\% of California had a canopy cover $\geq$ 5 m in 2020 ($\sim$ 81.826 km$^2$), highlighting the importance of forests in this region. The median canopy height of California was 11 m. This low median canopy height could reflect the 50\% decline in large trees ($\geq$ 61 cm diameter at breast height (dbh)) observed in California forests between the 1930s and the 2000s, a period during which large trees were replaced by smaller tree species with the goal of achieving higher tree density \citep{mcintyre2015,herbert2022}.

Regarding tall trees, 0.7\% of the forest shows trees with a median height above 40 m, and 0.07\% have a median height $\geq$ 50 m. Alongside our model, only the Sentinel-2-based height map \citep{lang2022} was able to estimate canopy height above 40 m (but not the individual trees). Having a map of these tall trees is important as they dominate the carbon stocks. For example, in the nearby Cascade Mountains crest in Oregon and Washington states, it has been shown that large trees, while accounting for 2 to 4\% of the stems, hold 33 to 46\% of the total carbon stored \citep{mildrexler2020}. Furthermore, in a sample of 48 forest sites worldwide, it has been observed that trees with a diameter at breast height (dbh) $\geq$ 60 cm comprised 41\% of aboveground live tree biomass, and the largest 1\% of trees with a dbh $\geq$ 1 cm comprised 50\% of aboveground live biomass \citep{lutz2018}.

Tall forests of California are renowned for the presence of coastal redwoods (\textit{Sequoia sempervirens}) along the Pacific coastal range and giant sequoias (\textit{Sequoiadendron giganteum}) endemic to the Sierra Nevada Mountains. These endangered species are among the tallest and most massive tree species in the world, capable of living thousands of years. Their preservation is essential for maintaining biodiversity in forest ecosystems \citep{piirto2002ecological,Francis2019,enquist2020}. While most giant sequoia groves are localized, the same cannot be said for coastal redwoods \citep{piirto2002ecological,Francis2019}. Combined with species mapping from spectral characteristics, which has been shown to be possible and accurate for the coastal \textit{Sequoia sempervirens} on a smaller scale \citep{Francis2019}, all the individuals of the species could be mapped. Large trees, as part of the megabiota, are more susceptible to extinction, and changes in their abundance disproportionately impact ecosystem and Earth system processes, including biomass, carbon, nutrients, and fertility \citep{enquist2020}. Since our model can estimate heights above 40 m and individual trees are visible in the predicted CHM, it could be utilized to locate all these trees and forests of primary importance for conservation.

\subsection{Advances in height and canopy structure mapping}



In comparison to existing canopy height maps \citep{tolan2023sub,lang2022,potapov2021m}, our results represent a significant advancement as they accurately reproduce the 3D structure of individual trees from a nadir view, similar to LiDAR CHMs. This unexpected achievement was made possible by the CNN's ability to perform geometric operations and recover 3D information from 2D images. With the canopy height map, we can now access more precise tree characteristics, such as height and crown size, location,  directly from VHR images. This is particularly important in California, where tall trees are mostly found in mountainous and hilly areas, and often appear distorted in VHR images. The latest developments in the co-registration of VHR images aim to accurately register the ground \citep{kristollari2022}, but there is still no available method to register trees. Accessing the 3D dimension of predicted CHMs from a nadir view at different time periods may assist in the co-registration of VHR images.

The results of Tolan's model seem to indicate that 3D reconstruction of individual tree height is not achieved from satellite VHR images (see Fig. \ref{2D3D} and \citep{tolan2023sub}). The cause is yet to be determined, whether it is due to the model or the characteristics of the satellite images, such as less accurate geolocations and diverse view angles. Furthermore, Tolan's model utilized two 8-GPU Voltas in the unsupervised pretraining phase \citep{tolan2023sub}, making it inaccessible for most research groups. In contrast, Our model employs the U-Net architecture, which runs on a single Nvidia RTX3090 GPU (24GB memory), and training/prediction can be made on a local machine. In a future experiment, we will assess if our model maintains 3D tree structure reconstruction when applied to VHR satellite images.

\subsection{Limitations}

Like all deep learning models, our model is subject to sampling bias. It relies on LiDAR data that is more prevalent in forested areas, limiting validation in regions with lower vegetation, like Chaparral, which is common in the southern Californian landscape. The model's performance is optimized for specific vegetation periods (April to August) and specific acquisition geometries. The performance of the model is not guaranteed with alternative configurations. To address misclassification of buildings as vegetation in reference CHMs, additional background data from a pre-existing building footprints dataset was incorporated. However, data availability for the additional background dataset may vary across regions.

\section{Conclusion}
In this work, we present the canopy height map of California at a sub-meter resolution (0.6 m) for the year 2020. We trained a deep learning regression model with the popular CNN architecture U-Net using aerial RGB-NIR NAIP images as input and LiDAR data as reference for canopy height. We demonstrate that in California, our model outperforms all existing remote-sensing derived canopy height maps. The observed 3D reconstruction of the tree structure from a VHR image has never been achieved before and could be used to gather individual information about the tree, such as height and crown size, or to produce maps of individual trees. The next steps are (i) to produce tree height data over the continental US with this method, and (ii) to apply this method to VHR images from satellites to see if 3D tree structure reconstruction is possible and can be used to map tree height on a global scale.

\section{Acknowledgements} 
The authors wish to thank the Grantham Foundation and High Tide Foundation for their generous gift to UCLA and support to \url{CTrees.org}. Part of this work was carried out at the Jet Propulsion Laboratory, California Institute of Technology, under a contract with the National Aeronautics and Space Administration (NASA).
Conceptualization, F.H.W., R.D, S.F. and S.S.; methodology, F.H.W., S.R., A.L.R., G.C., R.D., and S.F.; software, F.H.W., R.D., S.F. and M.C.M.H.; validation, F.H.W., S.R., and  A.L.R.; formal analysis, F.H.W. and S.R.; investigation, F.H.W., R.D., S.F., M.C.M.H. and S.S.; resources, S.S.; data curation, F.H.W. S.R., and  A.L.R.; writing—original draft preparation, F.H.W.; writing—review and editing, F.H.W., S.R., A.L.R., G.C., R.D., S.F., M.C.M.H., M.B., P.C. and S.S.; visualization, F.H.W. and M.C.M.H.; supervision, S.S.; project administration, S.S.; funding acquisition, S.S.

\section{References \label{bibby}}
\newcommand{\newblock}{}
\bibliographystyle{unsrtnat}  
\bibliography{references_height}

\newpage

\appendix
\renewcommand\thefigure{A.\arabic{figure}}    
\section{Supplementary figures}
\setcounter{figure}{0}    

 
 \begin{figure}[ht]
\centering
\includegraphics[width=0.9\linewidth]{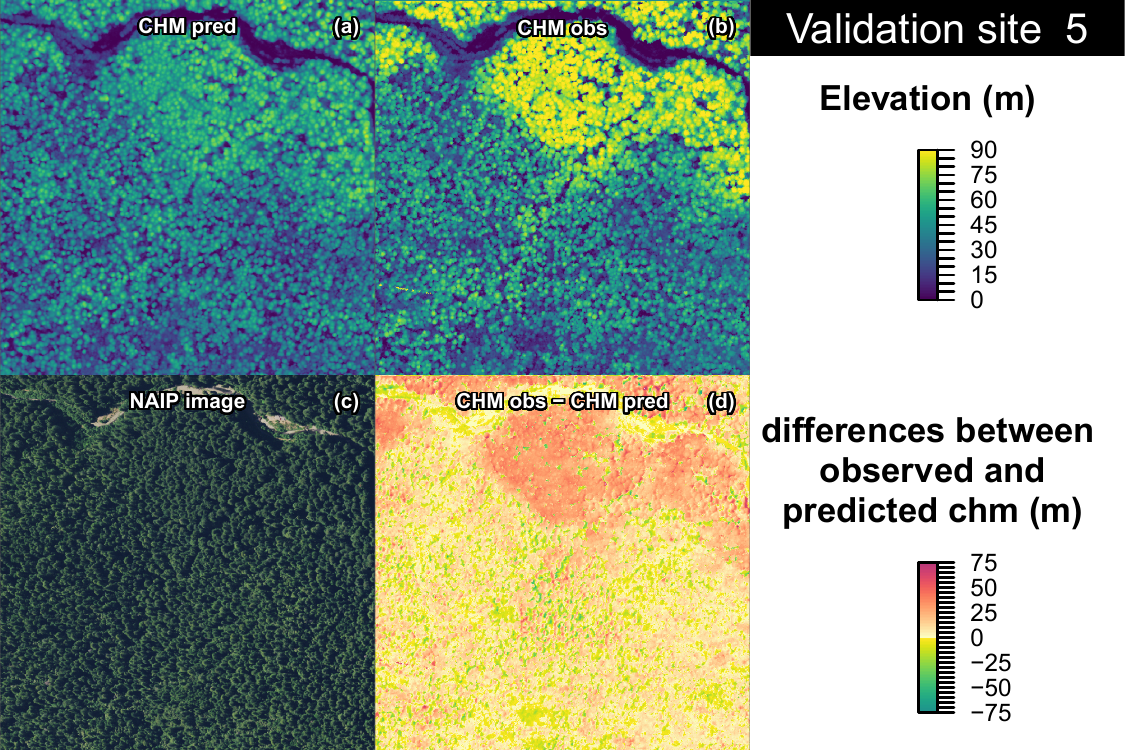}
 \caption{Predicted and observed canopy height, NAIP image (RGB), and the difference between observed and predicted Canopy Height Model for validation site 05.}
 \label{val05} 
 \end{figure}



\clearpage
 \begin{figure}[ht]
\centering
\includegraphics[width=0.9\linewidth]{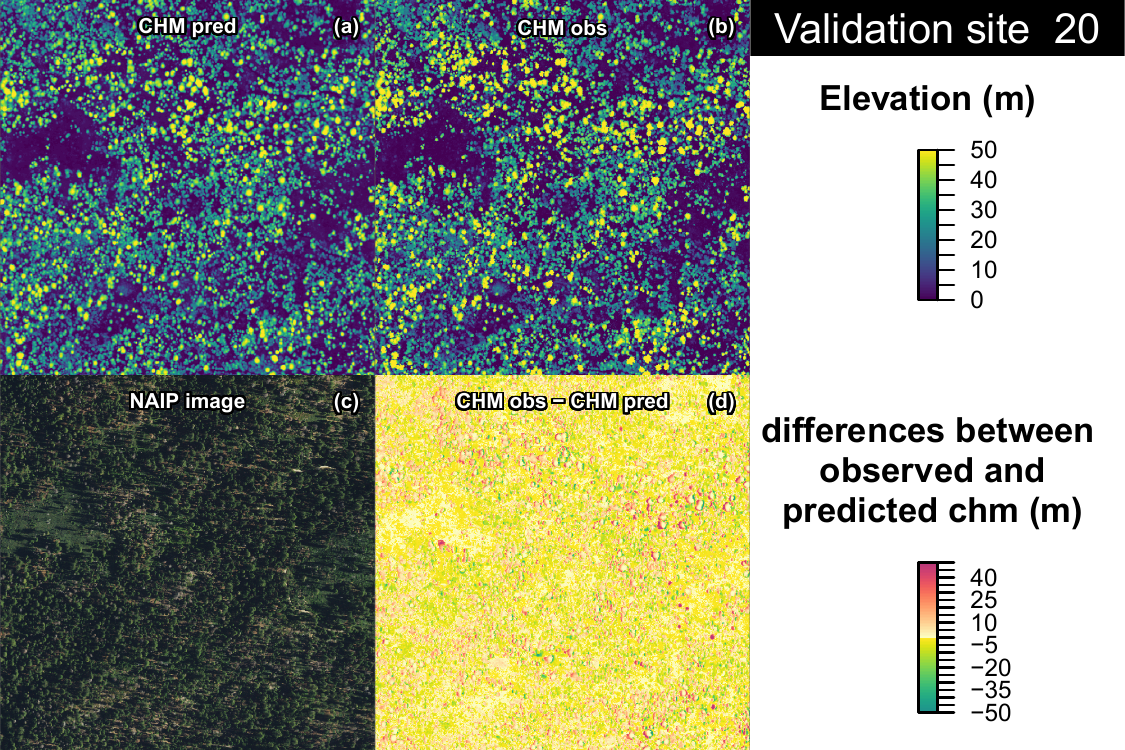}
 \caption{Predicted and observed canopy height, NAIP image (RGB), and the difference between observed and predicted Canopy Height Model for validation site 20.}
 \label{val20} 
 \end{figure}

  \begin{figure}[ht]
\centering
\includegraphics[width=0.9\linewidth]{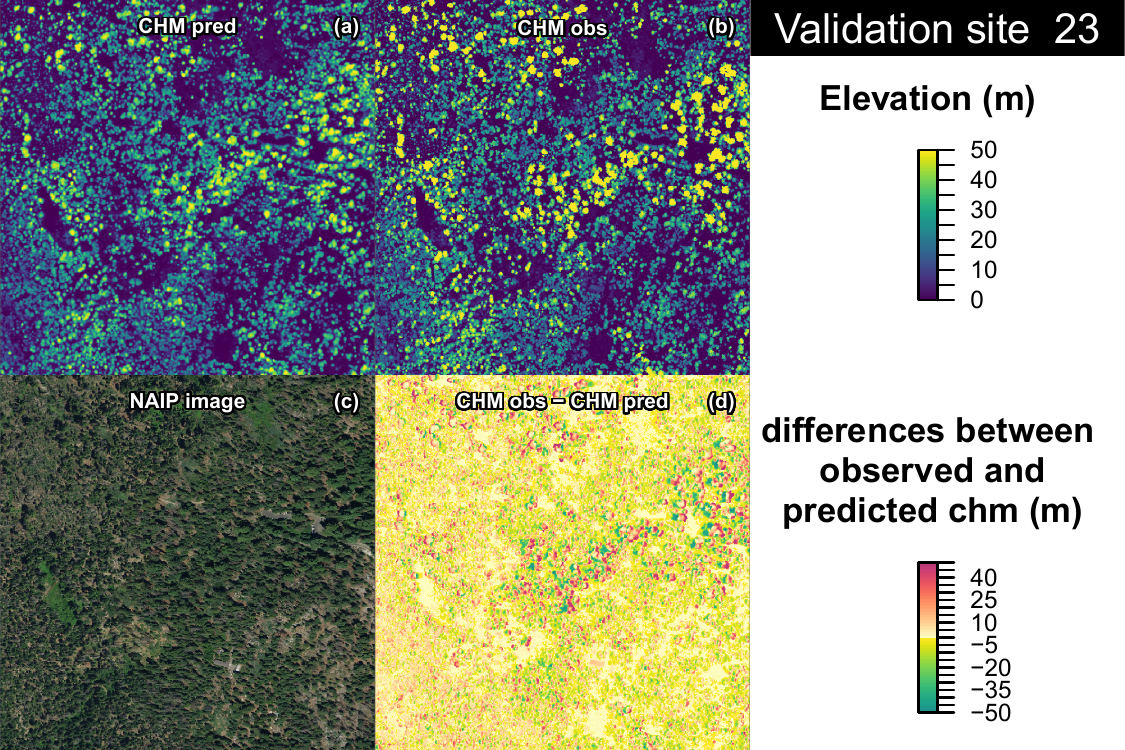}
 \caption{Predicted and observed canopy height, NAIP image (RGB), and the difference between observed and predicted Canopy Height Model for validation site 23.}
 \label{val23} 
 \end{figure}


\clearpage
 \begin{figure}[ht]
\centering
\includegraphics[width=0.90\linewidth]{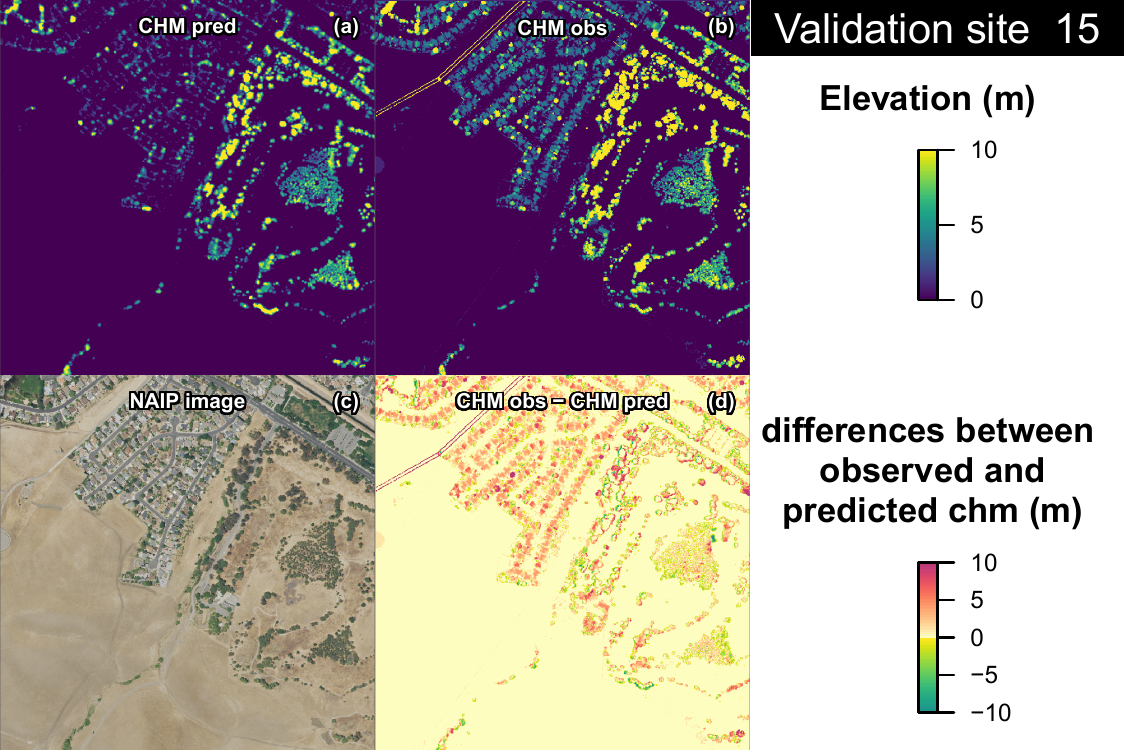}
 \caption{Predicted and observed canopy height, NAIP image (RGB), and the difference between observed and predicted Canopy Height Model for validation site 15.}
 \label{val15} 
 \end{figure}


 
  \begin{figure}[ht]
\centering
\includegraphics[width=0.9\linewidth]{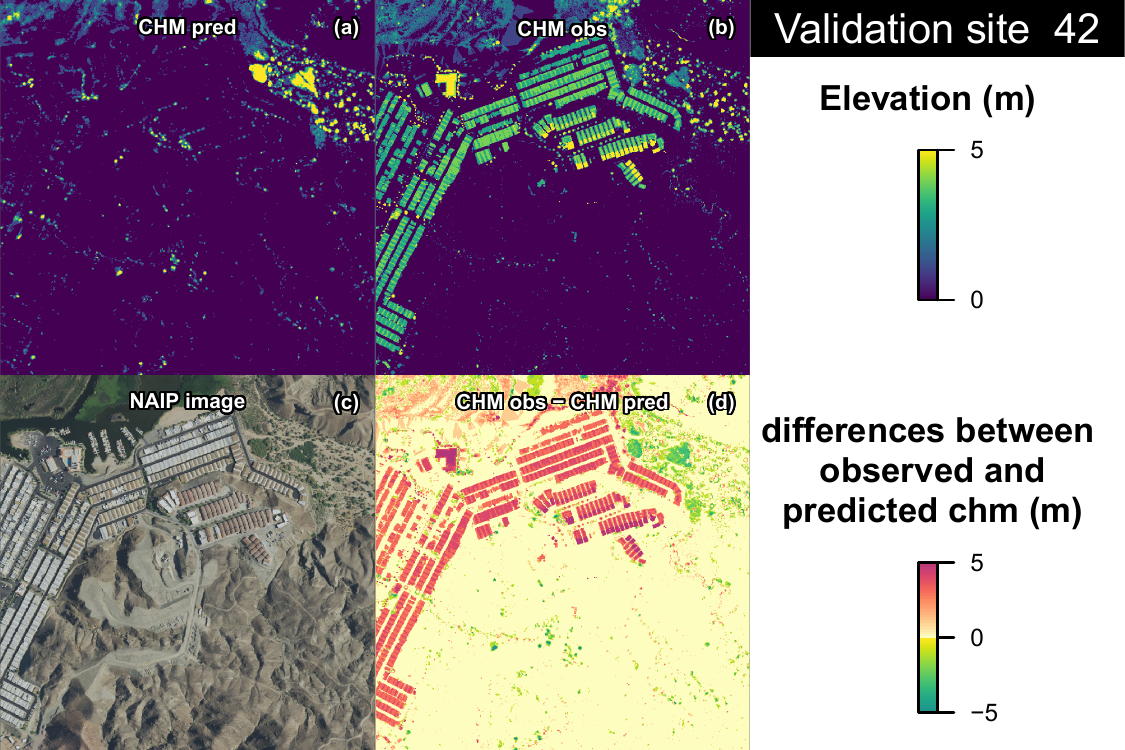}
 \caption{Predicted and observed canopy height, NAIP image (RGB), and the difference between observed and predicted Canopy Height Model for validation site 42.}
 \label{val42} 
 \end{figure}

 \clearpage
  \begin{figure}[ht]
\centering
\includegraphics[width=0.9\linewidth]{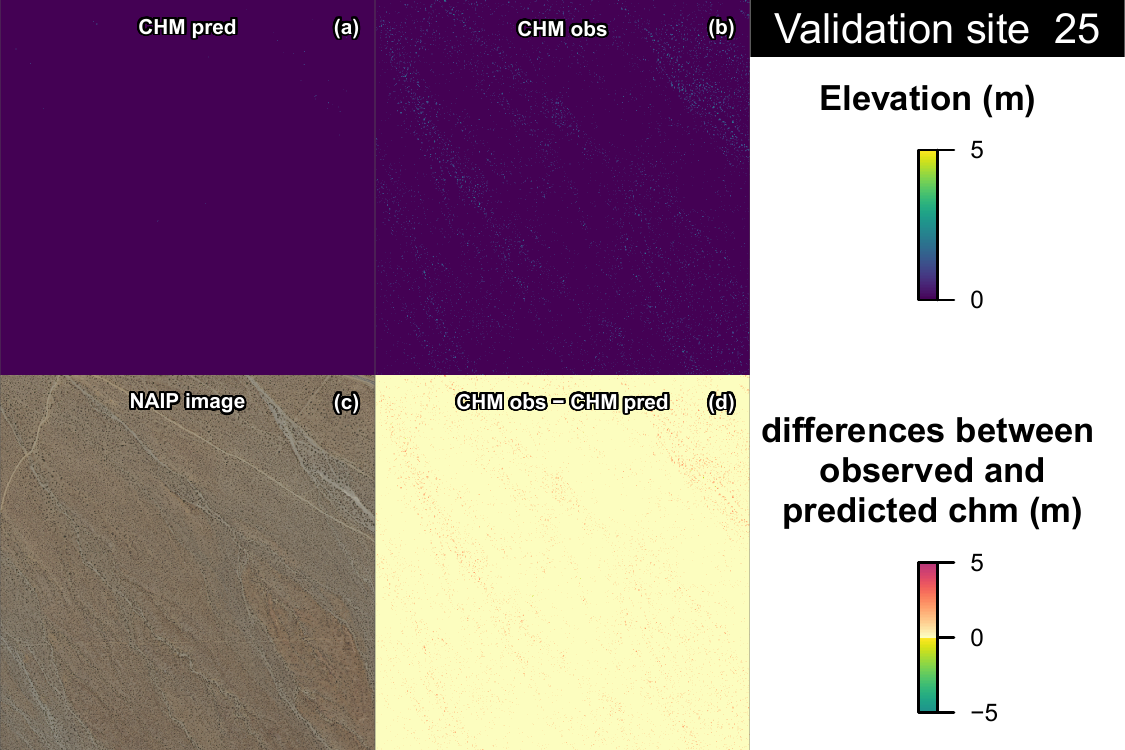}
 \caption{Predicted and observed canopy height, NAIP image (RGB), and the difference between observed and predicted Canopy Height Model for validation site 25.}
 \label{val25} 
 \end{figure}


\begin{figure}[ht]
\centering
\includegraphics[width=0.9\linewidth]{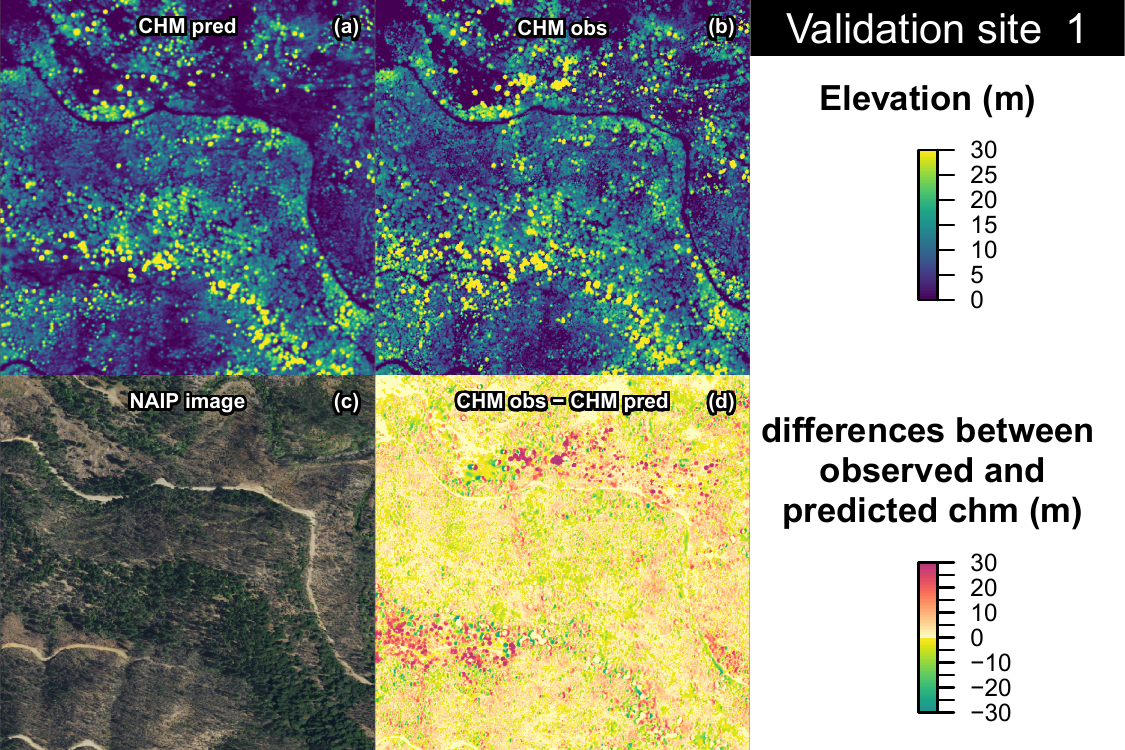}
 \caption{Predicted and observed canopy height, NAIP image (RGB), and the difference between observed and predicted Canopy Height Model for validation site 01.}
 \label{val01} 
 \end{figure}


\clearpage
\begin{figure}[ht]
\centering
\includegraphics[width=0.9\linewidth]{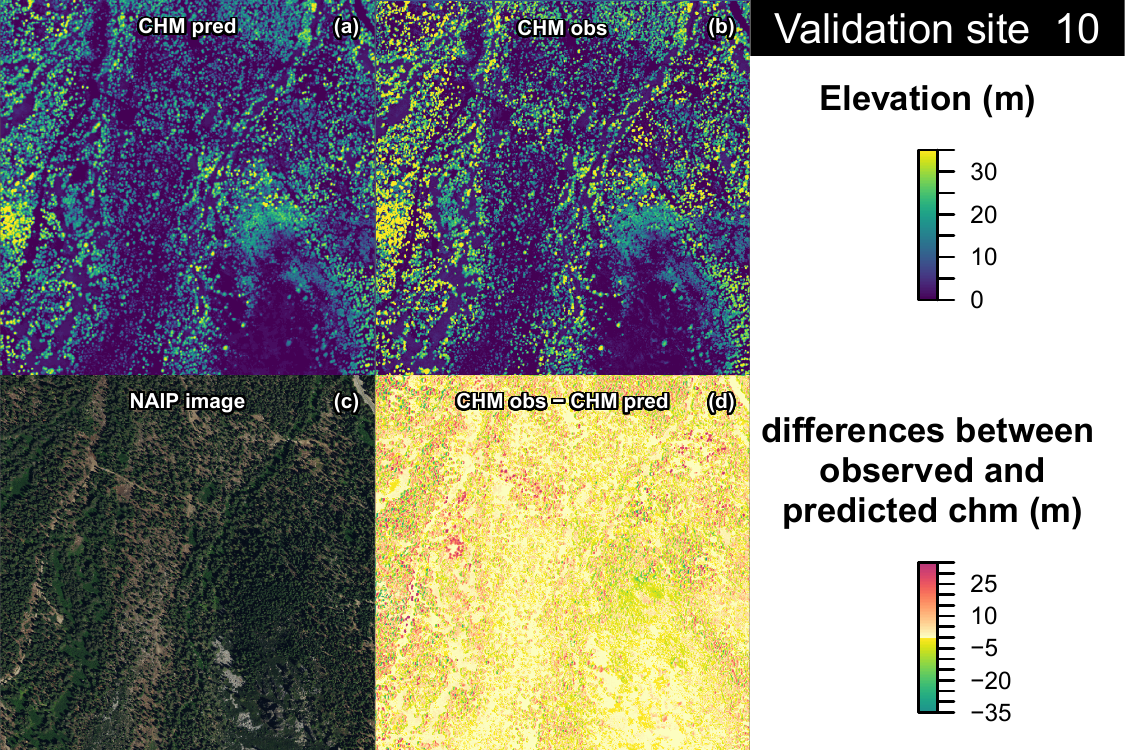}
 \caption{Predicted and observed canopy height, NAIP image (RGB), and the difference between observed and predicted Canopy Height Model for validation site 10.}
 \label{val10} 
 \end{figure}


\begin{figure}[ht]
\centering
\includegraphics[width=0.9\linewidth]{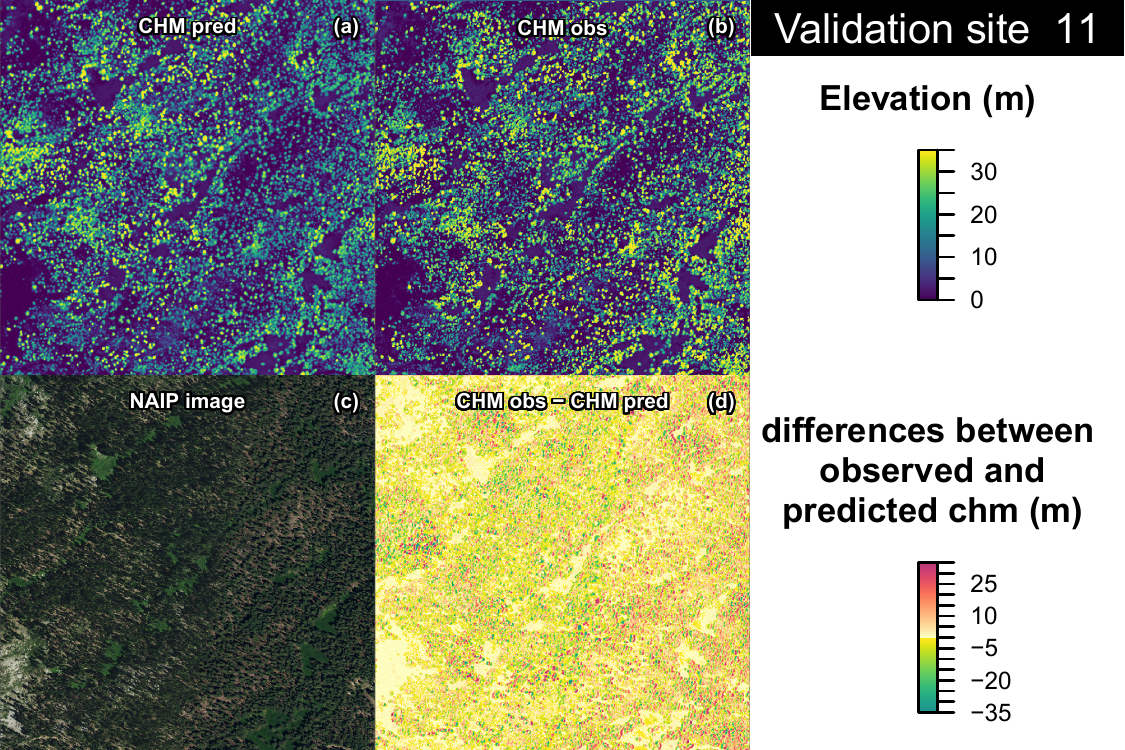}
 \caption{Predicted and observed canopy height, NAIP image (RGB), and the difference between observed and predicted Canopy Height Model for validation site 11.}
 \label{val11} 
 \end{figure}

\end{document}